\documentclass[10pt,twocolumn,letterpaper]{article}

\usepackage{cvpr}
\usepackage{times}
\usepackage{epsfig}
\usepackage{graphicx}
\usepackage{amsmath}
\usepackage{amssymb}
\usepackage{mathtools}
\usepackage{multirow}
\usepackage{enumitem}
\usepackage{color}
\def\red#1{\textcolor[rgb]{1,0,0}{#1}}
\def\blue#1{\textcolor[rgb]{0,0,1}{#1}}

\newcommand{\doublecheck}[1]{\textcolor{blue}{#1}}
\newcommand{\keypoint}[1]{\vspace{0.08cm}\noindent\textbf{#1}\quad}
\newcommand{\cut}[1]{}
\usepackage{dsfont}
\usepackage{algorithm}
\usepackage{algpseudocode}
\usepackage{makecell}
\usepackage{lipsum}
\usepackage{amsmath}
\usepackage{wrapfig}
\usepackage{bm}

\usepackage{multirow}

\algnewcommand\algorithmicforeach{\textbf{for each}}
\algdef{S}[FOR]{ForEach}[1]{\algorithmicforeach\ #1\ \algorithmicdo}

\makeatother
\usepackage[pagebackref=true,breaklinks=true,letterpaper=true,colorlinks,bookmarks=false]{hyperref}

\cvprfinalcopy 

\def\cvprPaperID{1273} 

\ifcvprfinal\fi
\begin{document}

\title{MetaHTR: Towards \emph{Writer-Adaptive} Handwritten Text Recognition}
\author{Ayan Kumar Bhunia\textsuperscript{1} \hspace{.2cm} Shuvozit Ghose\thanks{Interned with SketchX} \hspace{.2cm} Amandeep Kumar\footnotemark[1]\hspace{.2cm}   Pinaki Nath Chowdhury\textsuperscript{1,2}\hspace{.2cm}  \\ Aneeshan Sain\textsuperscript{1,2}\hspace{.2cm} Yi-Zhe Song\textsuperscript{1,2} \\
\textsuperscript{1}SketchX, CVSSP, University of Surrey, United Kingdom. 
\\
\textsuperscript{2}iFlyTek-Surrey Joint Research Centre
on Artificial Intelligence. 
\\{\tt\small \{a.bhunia, p.chowdhury, a.sain, y.song\}@surrey.ac.uk}.  \\ {\tt\small\{shuvozit.ghose,  kumar.amandeep015\}@gmail.com.}}

\maketitle
 
\ifcvprfinal\thispagestyle{empty}\fi

\begin{abstract}
\vspace{-0.2cm}
Handwritten Text Recognition (HTR) remains a challenging problem to date, largely due to the varying writing styles that exist amongst us. Prior works however generally operate with the assumption that there is a limited number of styles, most of which have already been captured by existing datasets. In this paper, we take a completely different perspective -- we work on the assumption that there is always a new style that is drastically different, and that we will only have very limited data during testing to perform adaptation. This \cut{results in} creates a commercially viable solution -- being exposed to the new style, the model has the best shot at adaptation, and the few-sample nature makes it practical to implement. We achieve this via a novel meta-learning framework which exploits additional new-writer data via a support set, and outputs a writer-adapted model via single gradient step update, all during inference (see Figure \ref{fig:Fig1_a.pdf}). We discover and leverage on the important insight that there exists few key characters per writer that exhibit relatively larger style discrepancies. For that, we additionally propose to meta-learn instance specific weights for a character-wise cross-entropy loss, which is specifically designed to work with the sequential nature of text data. Our writer-adaptive MetaHTR framework can be easily implemented on the top of most state-of-the-art HTR models. Experiments show an average performance gain of 5-7$\%$ can be obtained by observing very few new style data ($\leq$ 16). 

\end{abstract}


\vspace{-0.6cm}
\section{Introduction}\label{sec:intro}

Handwritten Text Recognition (HTR) has been a long-standing research problem in computer vision \cite{bhunia2019handwriting, poznanski2016cnn, wang2020decoupled, luo2020learn}. As a fundamental means of communication, handwritten text can appear in a variety of forms such as memos, whiteboards, handwritten notes, stylus-input, postal automation, reading aid for visually handicapped, etc  \cite{yousef2020origaminet}. In general, the target of automatic HTR is to transcribe handwritten text to its digital content \cite{shi2018aster} so that the textual content can be made freely accessible.
 
\begin{figure}[t]
\begin{center}
  \includegraphics[width=1\linewidth]{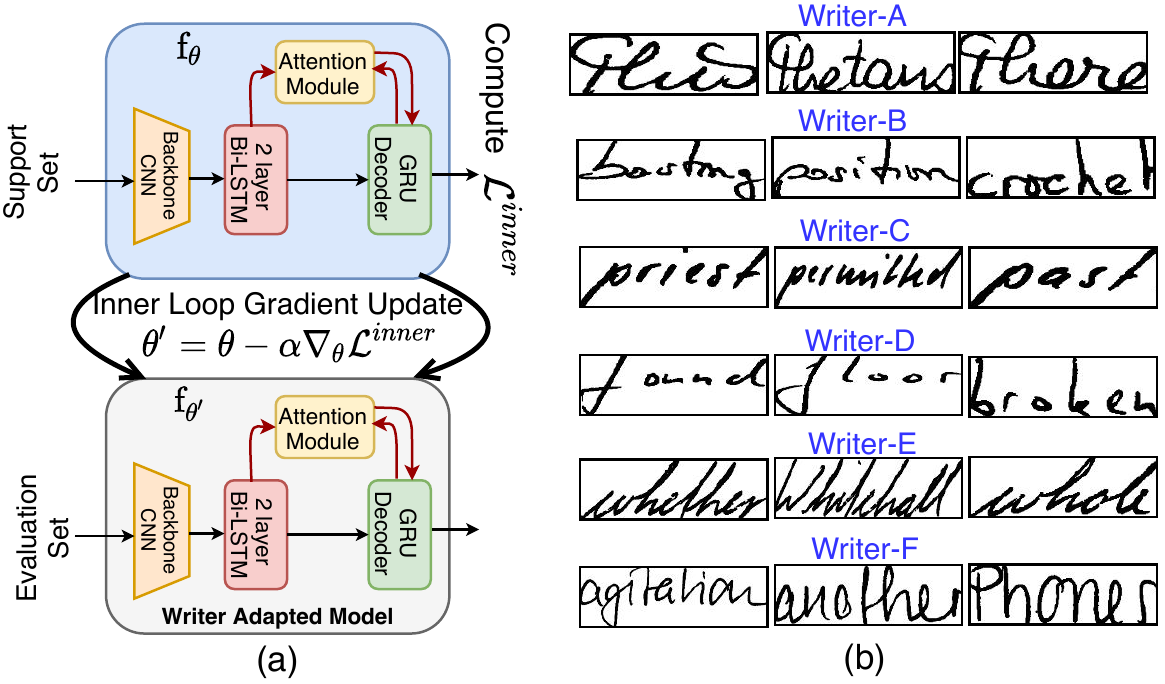}
\end{center}
\vspace{-0.40cm}
  \caption{
  (a) During inference, our MetaHTR framework exploits additional handwritten images of a specific writer through support set, and gives rise to a writer adapted model via single gradient step update. (b) Varying styles across different writers (IAM). 
  }
\vspace{-0.70cm}
\label{fig:Fig1_a.pdf}
\end{figure}

Handwriting recognition is inherently difficult due to its free flowing nature and \cut{the} complex shapes assumed by characters and their combinations~\cite{bhunia2019handwriting}. Torn pages, and warped or touching lines also make HTR more challenging. Most importantly however, handwritten texts are diverse across individual handwritten styles where each style can be very unique \cite{davis2020text, kang2020ganwriting} -- while some might prefer an idiosyncratic style of writing certain characters like `G' and `Z', others may choose a cursive style with uneven line-spacing.  


Modern deep learning based HTR models \cite{litman2020scatter, shi2018aster, li2019show} mostly tackle these challenges by resourcing to a large amount of training data. The hope is that most style variations would have already been captured because of the data volume. Albeit with limited success, it had become apparent that these models tend to over-fit to styles already captured, but generalising poorly to those unseen. This is largely owning to the \textit{uniquely} different styles amongst writers -- \textit{there is always a new style that is unobserved, and is drastically different to the already captured} (see Figure \ref{fig:Fig1_a.pdf}). The practical implication of this is, e.g., my iPad does not recognise my handwriting as well as it does for my 4-year-old. Our ultimate vision is therefore to offer an ``adapt to my writing'' button, where one is asked to write a specific sentence, so to make recognition performance of my own writing on par with that of my child.

Prior work on resolving the style gap remains very limited. A very recent attempt turns to training using synthetic data, so to help the model to become more accommodating towards new styles \cite{kang2020unsupervised}. However, synthetic data can hardly mimic all writer-specific styles found in the real-world, especially when the style is very unique. Although domain adaptation and generalisation approaches might sound viable, they generally do not offer satisfactory performance (as shown later in experiments), and require additional training via multiple gradient update steps. The sub-optimal performance can be mostly attributed to the large and often very unique domain gaps new writing styles bring, as opposed to the common dataset biases studied by domain adaption/generalisation. 




In this paper, we turn to a meta-learning formulation, which not only yields performances that are of potential {commercial} value (from 81.3\% to 89.2\% Word Recognition Accuracy), but also offers quick adaption (with just a single gradient update) using very few samples ($\le$16). 
The general motivation behind meta-learning \cite{finn2017model, nichol2018first, sun2019meta} matches ours very well -- absorbing information from related tasks and generalise onto unseen ones, by performing quick adaptation using a small set of examples \textit{during testing}. However, getting it {to} work with HTR has its own challenges, and to our knowledge has not been tackled before in the literature. The main challenges come from the inherent character sequence recognition nature of HTR, which is different to conventional meta-learning whose objective is mostly few-shot classification \cite{finn2017model, snell2017prototypical}. Furthermore, we importantly {discovers} that there also exists character-level style discrepancies, which when unaccounted for would trigger significant performance drop (see Section \ref{subsec:l2l}). 

To address these specific challenges, we first introduce a character-wise cross-entropy loss to our meta-learning framework. This albeit being a simple change, is crucial in light of the sequence recognition nature of our problem. We further guide the adaptation by introducing \emph{instance-specific weights} on top of the character-wise loss, instead of treating all characters equally by simply averaging \cite{shi2018aster, litman2020scatter}.  Modelling such character-specific weight is however non-trivial, as no fixed weight labels exist to supervise the learning process. Consequently, we let the model \textit{learning-to-learn} instance-specific weights for character-wise cross-entropy loss during the adaptation step. Our final model, MetaHTR, is therefore a meta-learning pipeline for writer-adaptive HTR where the model itself adaptively weights different characters to prioritise learning from more discrepant characters. {That is, during inference, our MetaHTR framework exploits few additional handwritten images of a specific writer through a support set, and gives rise to a writer-adapted model via a single gradient update step (see Figure \ref{fig:Fig1_a.pdf}).} Our meta-learning design can be coupled with any state-of-the-art HTR model, and empirical investigation shows that model agnostic meta-learning (MAML) pipeline \cite{finn2017model} provides a legitimate choice to design our MetaHTR framework upon.

Contributions of this paper can be summarised as follows: (1) We introduce for the first time, the problem of writer-adaptive HTR, where the model adapts to new writing styles with only very few samples during inference, (2) We introduce a meta-learning framework to tackle this new problem, by introducing learnable instance-wise weights for a character-specific loss specifically designed for HTR.  (3) We confirm that our framework consistently improves upon even the most recent state-of-the-art methods.

\vspace{-0.1cm}
\section{Related Works}
\vspace{-0.1cm}

\noindent \textbf{Text Recognition:} \cut{With the inception of deep learning, Jaderberg \etal  \cite{JaderbergARXIV2014, jaderberg2014deep} introduced a text recognition framework employing deep networks, which was restricted to dictionary words only. It was extended to unconstrained lexicon-free framework \cite{Jaderberg2015}, but was still limited by character level localisation for training. Poznanski \etal \cite{poznanski2016cnn} addressed  a CNN to estimate an n-gram frequency profile for HTR.} Connectionist Temporal Classification (CTC) layer \cite{GravesICML2006} made end-to-end sequence discriminative learning possible. Subsequently, CTC module was replaced by attention-based decoding mechanism \cite{lee2016recursive, shi2018aster} that encapsulates language modeling, weakly supervised character detection and  character  recognition  under a single model. This involves a rectification network to handle irregular text, followed by the final text recognition network. Needless to say attentional decoder became the state-of-the-art paradigm for text recognition for both scene text \cite{litman2020scatter, yu2020towards, yang2019symmetry, VeriSimilarECCV2018} and handwriting \cite{bhunia2019handwriting, luo2020learn, wang2020decoupled, zhang2019cvprDAtext}. Different incremental propositions have been made in this context, such as \cut{improving the rectification module \cite{VeriSimilarECCV2018, yang2019symmetry},} designing multi-directional convolutional feature extractor \cite{cheng2018aon}, improving attention \cite{cheng2017focusing, li2019show} and stacking multiple BLSTM layers for better context modelling \cite{litman2020scatter}.   

Besides word recognition accuracy\cut{only}, some works have focused on improving performance in low data regime, by designing adversarial feature deformation module \cite{bhunia2019handwriting}, learning optimal augmentation strategy \cite{luo2020learn}, and learning from synthetic training data via domain adaptation \cite{zhang2019cvprDAtext}. In this work, we introduce a new dimension of handwritten text recognition where model could be adapted during inference based on few handwritten word samples of the new writer in order to cope up with writer specific handwriting style.

\keypoint{Dealing with Writer Specificity:}Writer identification \cite{he2020fragnet, he2019deep} has been a long standing problem in the handwriting analysis community. Furthermore, the phenomenon of writer specific nature of handwriting is accepted in forensic science \cite{chen2019semi, he2020fragnet}, and handwritten signature \cite{hafemann2017learning} is used as an authentication medium in various official and banking sectors. Few shot writer-specific handwriting generation started in the online handwritten data \cite{aksan2018deepwriting} with coordinates, and has further been realised for offline handwritten images \cite{kang2020ganwriting}. Although the idea of style specific adaptation  \cite{platt1997constructive, gilloux1994writer, connell2002writer} has been introduced two decades ago, it has  been limited to online handwritten characters \cite{platt1997constructive}, handcrafted feature normalisation \cite{platt1997constructive}, few pre-defined handwritten styles (not user specific) and fixed lexicon-vocabulary \cite{connell2002writer}. Nevertheless, there has been no work encasing the full potential of end-to-end trainable deep models for writer adaptive lexicon free offline handwritten word recognition. Adaptation could be done without any increase of model parameters --leading to cost-effective deployment.

\keypoint{Meta-learning:} Meta-Learning aims to train a model on a series of related tasks, such that it learns the unseen task with only a few training samples~\cite{finn2017model}. One way is to\cut{to understand meta-learning is to} learn optimal initialization, such that it quickly adapts to new tasks with a few data \cite{snell2017prototypical,vinyals2016matching}. Various meta-learning algorithms\cut{have been proposed which} can be broadly categorised in three groups. While memory network based methods \cite{oreshkin2018tadam} learn across the task knowledge and aim to generalise to the unseen task, metric-based methods \cite{snell2017prototypical} aim to model a metric space in which learning is efficient with only a few samples.
The earlier two approaches are mostly architecture specific, and have been employed for few-shot classification problem. In contrast, there has been a significant attention towards using optimization based meta-learning algorithms \cite{finn2017model, nichol2018first, sun2019meta, antoniou2018train} due to its model-agnostic nature. Specifically, we choose the recently introduced algorithm, model-agnostic meta-learning (MAML) \cite{finn2017model} as it is compatible with any model which is trained with gradient descent and applicable to a variety of different learning problems. MAML aims to encode the prior knowledge into optimization process for fast adaption, and several variants have been proposed \cite{nichol2018first, antoniou2018train, rusu2018meta, stylemeup}. 
Later MAML++  \cite{antoniou2018train} introduced the sets of tricks to stabilize the training of the MAML. MetaSGD  \cite{li2017meta} proposes to train learnable learning rates for every parameter. Inspired by the success of  domain adaptive dialog generation \cite{qian2019domain}, 
we introduce MAML for writer adaptive handwritten text recognition. Nevertheless, we extend MAML for instant-adaptive sequence-recognition task over its off-the-shelf version \cite{finn2017model} that was initially proposed for non-sequential few-shot classification problem. 


\vspace{-0.2cm}

\begin{figure*}[t]
\begin{center}\label{fig:Fig2}
\includegraphics[width=1\linewidth]{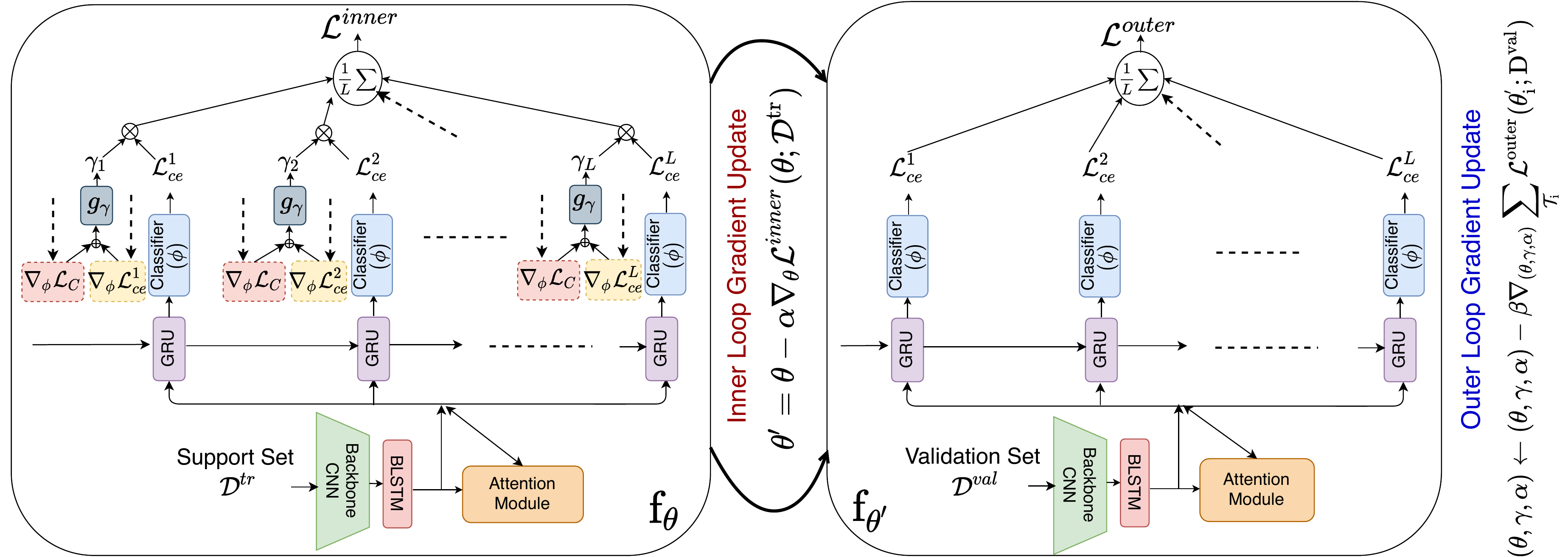} 
\end{center}
\vspace{-.15in}
  \caption{Our MetaHTR framework involves a bi-level optimisation process.  The \emph{ inner loop optimisation} computes learnable character instance-weighted loss $\mathcal{L}^{inner}$ upon the support set, followed by obtaining a pseudo-updated model ($\theta^{'}$). This includes a learnable character instance specific weight prediction module $(g_{\gamma})$ and learnable layer-wise learning rate parameters ($\alpha$). We expect $\theta^{'}$ to generalise  well on remaining validation set, thus finally updating the meta-parameters ($\theta, \gamma, \alpha$) by \emph{outer-loop} loss  $\mathcal{L}^{outer}$ over the validation set.}
\vspace{-.5cm}
\end{figure*}

\section{Methodology}

\vspace{-0.1cm}
 
\keypoint{Overview:}Traditionally, HTR model inputs a handwritten text image $X$ and generates output character sequence $Y = ( y_{1}, y_{2}, \dots, y_{L} )$, where L is the variable length of text. Conventional HTR models \cite{baek2019wrong} learn from multiple data instances often denoted as training dataset $\mathcal{D} = \{ (X_1, Y_1), (X_2, Y_2), \cdots, (X_N, Y_N)\}$. Due to data instance specific training, it ignores the writer specific data distribution, without modeling the shared common knowledge  \cite{hospedales2020meta} across different writers. Henceforth, the performance deteriorates on handwritten text images of unseen writers because of poor generalisation on diverse handwriting styles. 

In contrast, we take a meta-learning approach which seeks to learn the general rules of handwritten recognition from distribution of  multiple writer specific handwritten text recognition tasks. Let $W_S$ and $W_T$ denote the disjoint training and testing writer set respectively, i.e., $W_S \cap W_T = \O $. The training and testing sets are denoted as $\mathcal{D}^{S}=\{ \mathcal{D}_1^{S}, \cdots, \mathcal{D}_{|W_S|}^{S} \}$ such that $|W_S| > 1$ and $\mathcal{D}^{T} = \{ \mathcal{D}_1^{T}, \cdots, \mathcal{D}_{|W_T|}^{T}\}$ such that $|W_T| \geq 1$. Every  i$^{th}$ writer in both training and testing set, has its own set of $N_i$ labelled images as $\mathcal{D}_{i} = \{ (X_1, Y_1), (X_2, Y_2), \cdots, (X_{N_i}, Y_{N_i}) \}$. During training, data is sampled across writer specific tasks from training set $\mathcal{D}^{S}$ to learn a good initialization point $\theta$, by modeling the shared knowledge across different writers -- such that it can quickly adapt to any new writer using few examples. During inference, with respect to j$^{th}$ writer from testing set as $\mathcal{D}_j^{T}$, we consider to have access to $k$ (very few) labelled samples, based on which we update $\theta \mapsto \theta_j$ using just one gradient step to obtain a writer-specialised HTR model -- this is called $k$-shot adaptation.

\subsection{Baseline HTR Models} \label{basemodel_HTR}

Nowadays, state-of-the-art text recognition networks, many of which were originally proposed  for scene text \cite{shi2018aster}, are now simultaneously validated \cite{bhunia2019handwriting, luo2020learn, zhang2019cvprDAtext, wang2020decoupled} over HTR datasets, as both follows a unified framework and objective. Therefore, attentional decoder based pipeline being the current state-of-the-art for text recognition, we select three seminal works, namely ASTER \cite{shi2018aster}, SAR \cite{li2019show} and SCATTER \cite{litman2020scatter}, to use as our baseline HTR models. Moreover, ours is a meta-framework and could be adopted with most deep-text recognition pipelines.

For completeness, we briefly summarise the outline of text recognition model. In general, they consist of four components: (a) a convolutional feature extractor, (b) BLSTM layers for context modeling (c) a RNN decoder predicting the characters autoregressively one at a time step, and (d) an attentional block.  Let the extracted convolutional feature map be $\mathcal{F} \in \mathbb{R}^{h'\times w' \times d}$ for a rectified image input, where $h'$, $w'$ and $d$ signify height, width and number of channels. Every $d$ dimensional feature at $\mathcal{F}_{i,j}$ encodes a particular local image region based on the receptive fields, which can be reshaped into list of vectors $\mathrm{F} = [f_1, f_2, \cdots, f_Q ]$, where $Q = h' \times w'$. Thereafter BLSTM is employed to capture the long range dependencies on every position, thus alleviating the constraints of limited receptive field giving list of context rich vectors as: $\mathrm{H} = [h_{1}, h_{2}, ..., h_{Q}] =$ \texttt{BLSTM}$([f_{1}, f_{2}, \cdots, f_{Q}])$. At every time step t, the decoder RNN predicts an output character or end-of-sequence (EOS) $y_{t}$ based on three factors: a) previous internal state $s_{t-1}$ of decoder RNN, (b) the character $y_{t-1}$ predicted in the last step, and (c) a glimpse vector $g_t$ representing the most relevant part of $\mathcal{F}$ for predicting $y_{t}$. In order to get $g_t$, the previous hidden state $s_{t-1}$ acts as a query to discover the attentive regions as $g_t = \sum_{i=1}^{Q}a_{t}(i)\cdot h_i$. Attention score at $t^{th}$ time step $a_{t}(i)$ is computed by a function $\mathcal{A(\cdot)}$ as $\mathrm{a_{t}(i) = \sigma(v^{\top}tanh(W_{s}s_{t-1} + W_{h}h_{i} + b_{a})})$, $\sigma$ being softmax across spatial positions $i \in [1,Q]$, and $\{W_{s}, W_{h}, b_{a}\}$ are learnable weights. The current hidden state $s_{t}$ is updated by:  $\mathrm{{(o_{t}, s_{t}) =  RNN(s_{t-1}; \; [E(y_{t-1}), \; g_{t}] ) )}}$, where $E(.)$ is a character embedding layer with embedding dimension $\mathbb{R}^{128}$, and [.] signifies a concatenation operation. Finally, $\bm\tilde{y}_{t}$ is predicted as:
\vspace{-0.2cm}
\begin{equation}
p(\bm\tilde{y}_{t})=\mathrm{softmax(W_{o}o_{t} + b_{o})}  
\vspace{-0.2cm}
\end{equation}

 \noindent We denote the complete set of parameters for every baseline as $\theta$, and particularly that for the final classification layer as $\phi = \{W_o, b_o\}$.     SAR \cite{li2019show}  addresses 2D attention to eliminate the need of image rectification network \cite{shi2018aster}  and SCATTER \cite{litman2020scatter} couples multiple BLSTM layers for richer context modelling on the top of \cite{shi2018aster}. We refer  the reader to \cite{shi2018aster, li2019show, litman2020scatter} for further architectural details.

\subsection{Basics: Gradient Based Meta-Learning}\label{basics_maml}

A popular optimization-based meta-learning algorithm is model-agnostic meta-learning (MAML) \cite{finn2017model}. Here, the goal is to learn good initialization parameters $\theta$ that represent an across-task shared knowledge among related tasks, so that it can quickly adapt to any novel task of same distribution with only a few gradient update iterations.

Let $\mathcal{T}$ represent multiple tasks where $\mathcal{T}_i$ denotes the $i^{th}$ task sampled from some task distribution $p(\mathcal{T})$ i.e. $\mathcal{T}_i \sim p(\mathcal{T})$. In our case, $\mathcal{T}_i$  is sampled across a task containing labelled training data from a specific writer $\mathcal{D}_i^{S}$. Each task $\mathcal{T}_i$ consist of a support set $\mathrm{D}^{tr}$ and a validation set $\mathrm{D}^{val}$. Additionally, let a neural network be represented by $f_{\theta}$, where $\theta$ is the initial parameter of the network. Intuitively, MAML tries to find a good initialization of parameters $\theta$, representing the prior or meta-knowledge, so that a few updates of $\theta$ using $\mathrm{D}^{tr}$ can make large improvements by reducing the error measures and boosting the performance in $\mathrm{D}^{val}$. To learn this optimal initialisation parameter $\theta$, we first adapt (task-specific) $f_{\theta}$ to $\mathcal{T}_i$ using $\mathrm{D}^{tr}$ by fine-tuning:
\vspace{-.25cm}
\begin{equation}\label{eqn1}
    \theta^{'}= \theta - \alpha \nabla_\theta \mathcal{L}^{inner} (\theta; \mathrm{D}^{tr})
\vspace{-0.15cm}
\end{equation}
 
Evaluation of the adapted model is performed on unseen examples sampled from the same task $\mathrm{D}^{val} \in \mathcal{T}_i$, to measure the generalisation of $f_{\theta'}$. This acts as a feedback for MAML to adjust its initialization parameters $\theta$ to achieve better generalisation on any $\mathcal{T}_i$ (across-task):
\vspace{-.25cm}
\begin{equation}\label{eqn2}
    \theta \leftarrow \theta - \beta \nabla_\theta \sum_{\mathcal{T}_i} \mathcal{L}^{outer} (\theta^{'}; \mathrm{D}^{val})
\vspace{-0.15cm}
\end{equation}

\subsection{Meta-learning for Writer Adaptive HTR}\label{our_htr}
Let the baseline text recognition model in our case (parameterized by $\theta$) be represented as $f_\theta$. Instead of naive writer-specific fine-tuning that usually requires hundreds of gradient updates, we seek to learn the general rules \cite{finn2017model} of handwriting recognition using multiple writer specific handwritten text recognition tasks. Meta-training involves sampling tasks which here is defined with respect to each specific writer. In particular, $\mathcal{T}_i  \sim p(\mathcal{T})$ indicates selecting $\mathrm{2B}$ labelled samples from $i$-th writer training set $\mathcal{D}^{S}_i$, out of which we make $D^{tr}$ for inner loop update and $D^{val}$ for outer loop update each containing $B$ samples. It should be noted that  model parameters are updated by averaging gradients \cite{choi2020scene} of outer loop loss over a meta-batch $\{\mathcal{T}_1, \mathcal{T}_2, ..., \mathcal{T}_M\}$ having size of $M$. 

 \noindent \textbf{Character-Wise (CW) Loss for Meta-learning:} Given the backdrop of MAML (section \ref{basics_maml}) and any baseline text recognition model (section \ref{basemodel_HTR}), one can train a meta-learning model \cite{choi2020scene}, given an access to the loss function. A naive approach would be using traditional cross-entropy loss \cite{shi2018aster}, which usually trains any attentional-decoder based text recognition system, for both inner (Eqn. \ref{eqn1}) and outer loop (Eqn. \ref{eqn2}). If output from text-recognition model is $\bar{Y} = \{\bar{y_1}, \bar{y_2}, \cdots,  \bar{y_L}\}$, character-wise (CW) cross-entropy (ce) loss summed over the ground-truth output sequence $Y = \{y_1, y_2, \cdots, y_L\}$ can be defined as: 
 \vspace{-0.25cm} 
\begin{equation}\label{eqn_ce}
\mathcal{L}_{ce} =  -\frac{1}{L}\sum_{t=1}^{L} L_{ce}(y_t, \bar{y_t}) = -\frac{1}{L}\sum_{t=1}^{L} y_t \log p(\bar{y}_{t})
 \vspace{-0.1cm} 
\end{equation}

\subsection{Learning-to-Learn weights for CW Loss}
 \vspace{-0.1cm} 
\label{subsec:l2l}
 \noindent \textbf{Motivation:} Being a sequence recognition problem, $\mathcal{L}_C$ involves a summation operation \cite{shi2018aster} over the character sequence, thus treating every character specific cross-entropy loss equally. We conjecture this task specific adaptation for sequence recognition could be boosted if weight values for each character instance-specific loss are learned, such that the model adapts better with respect to those characters having a high discrepancy. 
 \begin{wrapfigure}{l}{0.22\textwidth}
    \vspace{-0.1cm}
    \centering
      \caption{Adaptation Analysis}
    \includegraphics[width=0.25\textwidth]{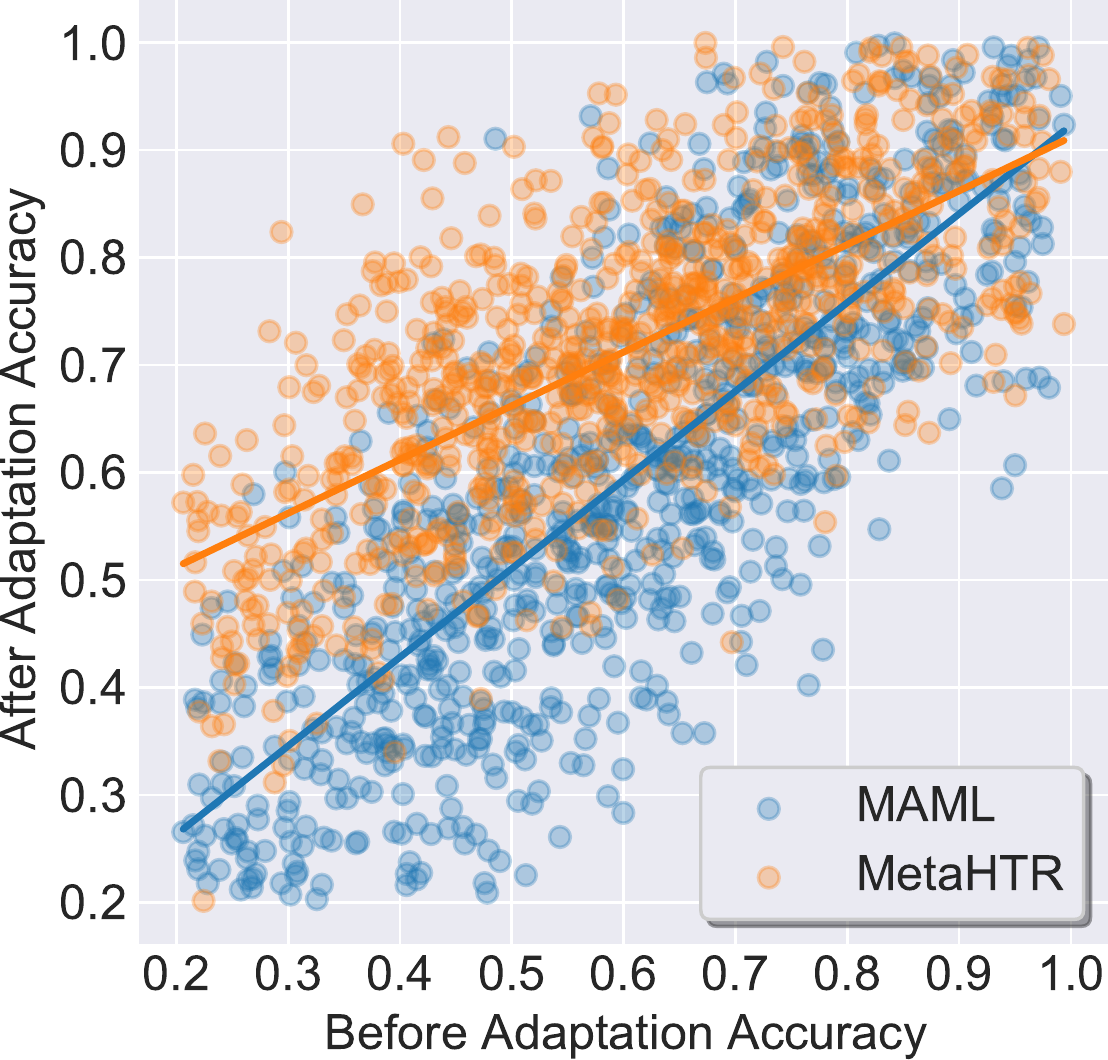}
    \label{fig:plot}
     \vspace{-0.3cm}
    \vspace{-0.5cm}
\end{wrapfigure}
Intuitively speaking, our model learns knowledge across tasks, where given a word `covid' from a new specific writer, properties of certain handwritten characters (e.g. `c', `v', `i') could be close to the encoded knowledge of MAML's initialisation parameter \cite{baik2020learning}, to enhance easier recognition. On the contrary, significant discrepancy could exist among certain characters (e.g. `o', `d') that are difficult to recognise using average knowledge encapsulated inside MAML's initialisation parameter. Thus, during fast adaptation, the model needs to update itself by prioritising the optimisation with respect to those particular characters  (e.g. `o', `d') whose style variation  is more towards unknown to the model's initialization. In other words, for faster adaptation via inner loop loss, we intend to learn the instance specific weight of character-wise cross-entropy loss instead of simply averaging over all characters. Recent literature shows that meta-learning provides the flexibility to learn any hyper-parameters \cite{finn2017model}, parameterized loss functions \cite{chebotar2019meta}, learning rates \cite{li2017meta}, or weight attenuation \cite{baik2020learning} in the meta-learning process itself.  {Character wise recognition accuracy from different writers before and after adaptation are plotted in Figure~\ref{fig:plot}. It can be seen that the characters getting low accuracy using MAML's initialisation parameter before adaptation (X axis) also get low accuracy even after the adaptation (Y axis). However, our proposed learnable character-instance specific weight of MetaHTR helps to enhance the performance of those discrepant characters after adaptation. More insightful analysis is in Section \ref{ablation}.}

\keypoint{Meta-Optimisation:} Naturally, the question arises what information could be used to determine these weights. Recent studies show that the gradients used for fast adaptation (inner loop) contains the information \cite{baik2020learning} related to disagreement (e.g. this knowledge further needs to be learned or accumulated in the adaptation process) with respect to model's initialization parameters. As calculating gradients with respect to all the model's parameters is quite cumbersome, we calculate gradient of $t$-th character specific cross-entropy loss with respect to final classification layer (parameter $\phi$) as $\nabla_{\phi} \mathcal{L}^{t}_{ce}(\theta)$. It is then concatenated with gradients of mean loss (Eqn. \ref{eqn_ce}) which sums over character sequence with respect to $\phi$ (both gradient matrix being flattened) as $\mathcal{G}_t = \texttt{concat}\big(\nabla_{\phi} \mathcal{L}^{t}_{ce}(\theta), \nabla_{\phi} \mathcal{L}_c(\theta)\big)$. We postulate that gradient of the mean and character-instance specific losses provide knowledge towards determining how to weigh different character specific losses. Thus, we pass this $\mathcal{G}_t$ through a network $g_\gamma$ predicting a scalar weight value for $t$-th character specific loss as:

\vspace{-0.25cm} 
\begin{equation}\label{eqn_gamma} 
\gamma_t = g_\gamma \big(\texttt{concat}\big(\nabla_{\phi} \mathcal{L}^{t}_{ce}(\theta), \nabla_{\phi} \mathcal{L}_c(\theta)\big)\big)
\vspace{-0.05cm}
\end{equation}

\noindent Here, $g_\gamma$ is a 3-layer MLP network of parameters $\gamma$ followed by a sigmoid to generate weights. Therefore, the instance weighted inner loop loss becomes:
\vspace{-0.25cm} 
\begin{equation}\label{eqn6} 
\mathcal{L}^{inner} = - \sum_{t=1}^{L} \gamma_t \cdot  L_{ce}(y_t, \bar{y_t})
\vspace{-0.25cm}
\end{equation}

Traditional MAML uses a predefined constant learning rate $\alpha$ in  the inner-level optimization. Inspired from \cite{li2017meta, wang2020tracking}, we specify a learnable rate for each layer as follows:
\vspace{-0.20cm}
 \begin{equation}\label{eqn6_1} 
    \theta^{'}= \theta - \mathbf{\alpha} \cdot \nabla_\theta \mathcal{L}^{inner}(\theta, D^{tr})
\vspace{-0.1cm}
\end{equation}
\noindent where $\mathbf{\alpha}$ is a vector of size equal to number of layers in the baseline HTR model. The outer-loop loss is kept as traditional $\mathcal{L}_C$ (see Figure \ref{fig:Fig2}). Please note that $\theta^{'}$ is dependant on $\{\theta, \gamma, \alpha\}$ through the inner loop update (Eqn. \ref{eqn6_1}), and  all three meta-parameters $(\theta, \gamma, \alpha)$ are meta-learned via the outer-loop update as  $(\theta, \gamma, \alpha) \leftarrow (\theta, \gamma, \alpha)  - \beta \nabla_{(\theta, \gamma, \alpha)}  \sum_{\mathcal{T}_i} \mathcal{L}^{outer} (\theta^{'}_{i} ; \mathrm{D}^{val})$. Training and inference process is summarised in Algorithm \ref{algo:train} and \ref{algo:infer}, respectively.
 
\setlength{\textfloatsep}{0pt}
\begin{algorithm}[!hbt]
\sloppy
	\caption{Training for Writer Adaptive MetaHTR}
	\label{algo:train}
	\begin{algorithmic}[1]
        \State \textbf{Input}: Training dataset $\mathcal{D}^{S}=\{\mathcal{D}^{S}_1, \mathcal{D}^{S}_2 \dots, \mathcal{D}^{S}_{|W_S|}\}$; $\beta$ as learning rate. 
        \State \textbf{Initialise}: Initialise $\theta, \gamma, \alpha$
        \State \textbf{Output}: Optimised meta-parameters $\{\theta, \gamma, \alpha\}$
        \While {not done}
            \State Sample writer specific $\mathcal{T}_i = \{\mathrm{D}^{tr}_i, \mathrm{D}^{val}_i \} \sim p(\mathcal{T})$
            \ForEach {task $\mathcal{T}_i$}
                \State Evaluate inner objective: $\mathcal{L}^{inner}(\theta; \mathrm{D}^{tr}_i)$
                \State Adapt: $\theta^{'}_i = \theta - \alpha \nabla_\theta \mathcal{L}^{inner}(\theta; \mathrm{D}^{tr}_i)$
                \State Compute outer objective: $\mathcal{L}^{outer}(\theta^{'}_i; \mathrm{D}^{val}_i)$
             \EndFor
             \State Update meta-parameters:  $(\theta, \gamma, \alpha) \leftarrow (\theta, \gamma, \alpha)  - \indent \beta \nabla_{(\theta, \gamma, \alpha)}  \sum_{\mathcal{T}_i} \mathcal{L}^{outer} (\theta^{'}_{i} ; \mathrm{D}^{val})$
        \EndWhile
	\end{algorithmic} 
\end{algorithm} 

\vspace{-0.0cm}
\begin{algorithm}[!hbt]
\sloppy
	\caption{Inference for Writer Adaptive MetaHTR}
	\label{algo:infer}
	\begin{algorithmic}[1] 
        \State \textbf{Input}: Testing dataset $\mathcal{D}^{T} = \{\mathcal{D}^{T}_1, \mathcal{D}^{T}_2, \dots, \mathcal{D}^{T}_{|W_T|}\}$; meta-learned model parameters $\{\theta, \gamma, \alpha \}$, number of gradient updates $n$, a given writer $j$. 
            \State Get the available support set $\mathrm{D}^{tr}_j \in \mathcal{T}_j$
            \For {$n$ steps}
                \State Evaluate inner objective: $\mathcal{L}^{inner}(\theta; \mathrm{D}^{tr}_j)$
                \State Adapt: $\theta^{'}_j= \theta - \alpha \nabla_\theta \mathcal{L}^{inner}(\theta; \mathrm{D}^{tr}_j)$
            \EndFor
            \State \textbf{Return} \emph{Writer specialised} HTR model params. $\mathbf{\theta^{'}_j}$.
	\end{algorithmic} 
\end{algorithm}

\vspace{-0.3cm}
\section{Experiments}\label{sec:experiments}
\vspace{-0.1cm}
\noindent \textbf{Datasets:} We evaluate the performance of our writer adaptive MetaHTR on two popular datasets of Latin scripts, IAM \cite{marti2002iam} and RIMES \cite{grosicki2009icdar}. While IAM contains a total number of 1,15,320 English handwritten word images written by 657 different writers, RIMES consists of 66,982 French word images of 1300 different writers. Both datasets contain word samples with annotated writer information, thus enabling sampling of writer specific meta-batches, to perform episodic training \cite{finn2017model}. For RIMES, we use samples from a subset of 375 writers which is usually used in writer identification task \cite{siddiqi2010text} as well. Following \cite{bhunia2019handwriting}, we use the same partition for training, validation and testing as provided for IAM, while the partition released by ICDAR 2011 competition is used for RIMES. 

\noindent \textbf{Implementation Details:} Following traditional supervised learning protocols \cite{baek2019wrong}, we first pre-train every considered baseline HTR models using ADADELTA optimiser with learning rate 1, and a batch size 64. Thereafter, we perform the meta-training process on pre-trained baseline model's parameters for 20 epochs according to Algorithm \ref{algo:train}. Only one inner loop update is used during inference (Algorithm \ref{algo:infer}) unless otherwise mentioned.  \cut{The reported performance of MetaHTR model uses only one inner loop update during inference (see  Algorithm \ref{algo:infer}).} Additionally, the effect of increasing inner loop updates is shown in our ablative study (section \ref{ablation}). During meta-training, we consider meta-batch size of $M=8$ -- our meta-batch comprises 8 different writer specific tasks $\mathcal{T}_i$, that are used for updating the meta-parameters by taking average gradient. Within each task, the batch-size of support and validation set is $B=16$. We use ADAM as meta-optimiser with outer-loop learning rate $\beta$ as 0.0001, while the inner-loop learning  $\alpha$ is meta-learned along with instance-specific weight $\gamma_t$ of character-wise cross-entropy loss. We  implemented  our framework in PyTorch  \cite{paszke2019pytorch} and conducted experiments on a 11 GB Nvidia RTX 2080-Ti GPU.  

\noindent \textbf{Evaluation Metric:} We use Word Recognition Accuracy (WRA)  \cite{bhunia2019handwriting} for both \textit{with}-Lexicon (\textbf{L}) and \textit{with} No-Lexicon (\textbf{NL}) (unconstrained) HTR. As there is no separate adaptation set (support set) explicitly defined for testing set writers $W_T$ in either of these datasets, we do the following \cut{for fair comparison}: let $N_j^T$ be the total number of images under test writer $j$, we take random $k$ images (for $k$-shot adaptation) as the support set for adaptation and the adapted model is evaluated on remaining $(N_j^T - k)$ images. We do this for ten times, and cite average result to reduce the randomness. 
We use $k=16$ for our cited results unless mentioned otherwise. Due to this adaptation set constraint, only those writers having more than $32$ word images, contribute towards accuracy calculation. For fairness, we ensured uniform adaptation and testing set for all the competitive baselines.

\vspace{-.1in}
\begin{figure}[!hbt]
\begin{center}
  \includegraphics[width=0.95\linewidth]{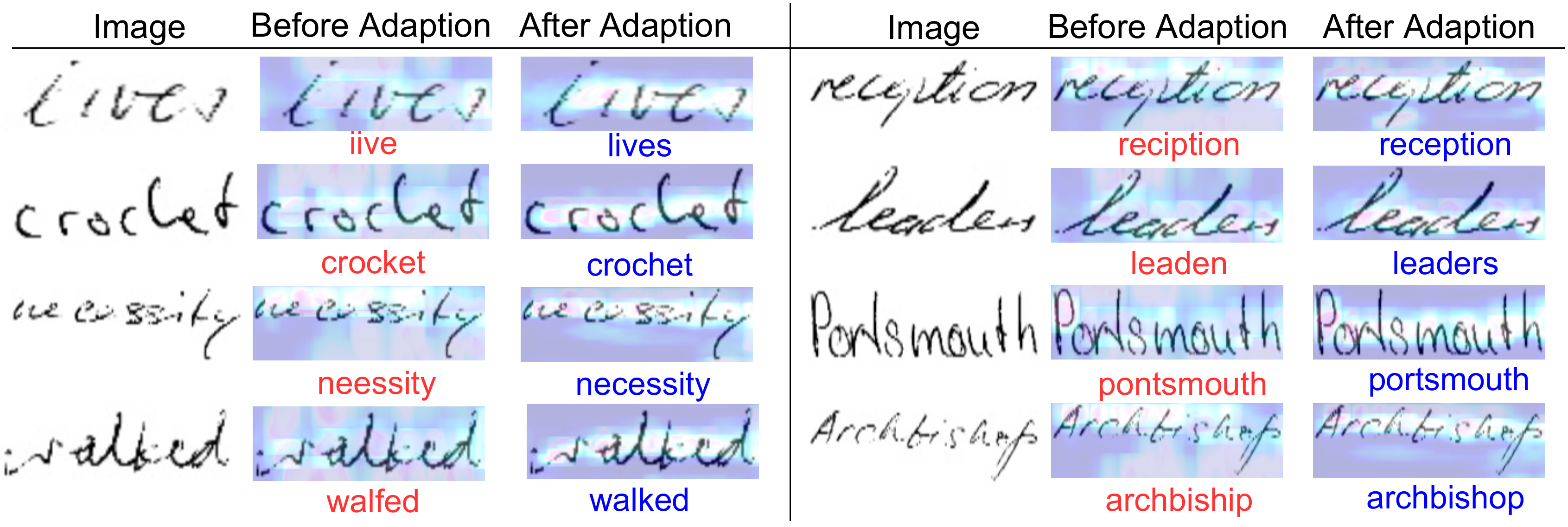}
\end{center}
\vspace{-.20in}
  \caption{Examples showing how adaptation leads to more consistent character-aligned attention map, followed by better recognition performance.  (\red{Red}: Incorrect, \doublecheck{Blue}: Correct)
}
\label{fig:fig3}
\end{figure}

\begin{table*}[!h]
\centering \label{main_table1}
\caption{Comparison among Baselines, naive Fine-tuning, and MetaHTR for using Lexicon (L), No Lexicon (NL). GAP: difference between MetaHTR (NL) vs Baseline (NL). We almost get around 5-7\% WRA improvement over respective baselines under NL setting.} 
\label{tab1}
\scriptsize
\begin{tabular}{c|cc|cc|ccc|cc|cc|ccc}
\Xhline{2\arrayrulewidth}
 &  \multicolumn{7}{c|}{{IAM} \cite{marti2002iam} (WRA)} & \multicolumn{7}{c}{{RIMES} \cite{augustin2006rimes} (WRA)}\\
 \cline{2-15}
Methods & \multicolumn{2}{c|} {{Baseline}} & \multicolumn{2}{c|} {{Fine-tuning}} &\multicolumn{3}{c|}{{MetaHTR}}&  \multicolumn{2}{c|}{{Baseline}} & \multicolumn{2}{c|} {{Fine-tuning}} & \multicolumn{3}{c}{{MetaHTR}}\\
\cline{2-15}
 & L & NL & L & NL & L & NL & GAP & L & NL & L & NL & L & NL & GAP \\
 \hline
    ASTER \cite{shi2018aster} & 90.3 & \blue{81.3} & 90.5 & 81.7 & 94.1 & \red{89.2} & \red{7.9}	$\uparrow$ & 93.6 & \blue{87.4} & 93.7 & 87.7 & 96.5 & \red{93.4} & \red{6.0}	$\uparrow$ \\
    \hline
    SAR \cite{li2019show}  & 91.6 & \blue{84.4} & 91.7 & 84.7 & 94.8 & \red{91.5} & \red{7.1}	$\uparrow$ & 93.8 & \blue{88.7} & 93.8 & 88.8 & 96.5 & \red{93.7} & \red{5.0} 	$\uparrow$\\
    \hline
    SCATTER \cite{litman2020scatter} & 91.7 & \blue{84.6} & 92.0 & 85.1 & 94.8 & \red{91.6} & \red{7.0}	$\uparrow$ & 93.8 & \blue{88.8} & 93.9 & 89.0 & 96.6 & \red{93.9} & \red{5.1}	$\uparrow$ \\
    \hline
\end{tabular}
\vspace{-0.5cm}
\end{table*}

\vspace{-.5cm}
\subsection{Competitors}\label{Competitors}
\vspace{-.10cm}
To the best of our knowledge, there exists no prior work particularly dealing writer specific adaptation for offline handwritten word images. However, we design several strong baselines from \emph{five} different perspectives to justify our MetaHTR framework. \textbf{\emph{(i)}  Learning Augmentation Approach:} Recently, there have been attempts to learn efficient data-augmentation strategy  to learn the style-variation present in the handwritten data using a learnable agent \cite{luo2020learn} or an adversarial feature deformation module  \cite{bhunia2019handwriting}. 
 \cut{Bhunia \etal \cite{bhunia2019handwriting} introduces a feature deformation module based on adversarial learning, while Luo \etal \cite{luo2020learn} uses  a learnable agent to discover the optimal text image augmentation strategy.} \textbf{\emph{(ii)}  Generative Approach:} One can synthetically generate \cite{kang2020ganwriting} multiple handwritten images with different words mimicking someone's handwriting style from few given handwritten examples (adaptation set). Thereafter, naive fine-tuning \cite{triantafillou2019meta} could be done over large (5K in our experiment) synthetically generated data considering them as writer's style specific training-set. \textbf{\emph{(iii)} Meta-Learning based Adaptation:} We follow this paradigm in our MetaHTR framework, however, there could be some off-the-shelf alternatives. An obvious choice could be naive-fine-tuning \cite{triantafillou2019meta} over the same labelled images of the adaptation set. We compare our method with typical MAML \cite{finn2017model} formulation, along with its first-order (MAML-FO) approximated version \cite{finn2017model} to judge how far the performance drops, while improving the computational speed. We compare with MetaSGD \cite{li2017meta} which uses learnable learning rate for each parameter and ANIL \cite{raghu2019rapid}, where only final classification layer ($\phi$) is pseudo-updated in the inner loop for computational ease. \textbf{\emph{(iv)} Domain Adaptation (DA) Approach:} All the training writers are considered as a single source domain, and the trained model is adapted using samples from a specific writer using adversarial learning as used in \cite{kang2020unsupervised}.
\textbf{\emph{(v)} Domain Generalisation (DG) Approach:} Domain generalisation aims to learn a generalised model via episodic training from writer-specific task distribution, which can directly perform well across unseen writers without any further gradient update. Following \cite{guo2020face}, we can twist our meta-learning pipeline to fit the objective of DG. Thus, we optimise the baseline HTR model using weighted ($\lambda=0.5$) summation for gradient (over meta-train set) and meta-gradient (over meta-test split through inner loop update). Mathematically, using our notation: $\operatorname*{argmin}_{\theta} \lambda \cdot \mathcal{L}(\theta; D^{tr}) + (1-\lambda) \cdot \mathcal{L}(\theta' ;D^{val})$, where $\mathcal{L}$ is the loss function (see Eqn. \ref{eqn_ce}) and $\theta'$ is pseudo-updated parameter by inner loop with learning rate 0.0005. It is worth noting that although DG \cite{guo2020face} and augmentation based approaches \cite{luo2020learn} cannot be compared directly to ours, as they do not involve any model-updation step at test time. \cut{but they still act as reasonable alternatives.} \cut{The two earlier approaches use same adaptation set as used in our MetaHTR.} 

 
\vspace{-0.2cm}
\begin{table}[]
  \tiny
    \centering
    \caption{Performance analysis with different approaches.}
    \resizebox{1\columnwidth}{!}{
    \begin{tabular}{cc|cc|cc}
        \Xhline{2\arrayrulewidth}
        \multirow{3}{*}{} & & \multicolumn{2}{c|}{IAM} & \multicolumn{2}{c}{RIMES} \\
        \cline{3-6}
         & & L & NL & L & NL \\
         \hline
         
        \multirow{3}{*}{\rotatebox[origin=c]{90}{DG}} & ASTER \cite{shi2018aster} + DG & 91.7 & 84.6 & 93.9 & 89.1 \\
         &  SAR \cite{li2019show} + DG & 92.4 & 85.7 & 94.3 & 89.5  \\
         &  SCATTER \cite{litman2020scatter} + DG  & 92.5 & 85.9 & 94.6 & 89.7 \\
         \hline
         
        \multirow{3}{*}{\rotatebox[origin=c]{90}{DA}} & ASTER \cite{shi2018aster} + DA & 90.3 & 81.1 & 93.8 & 87.4 \\
         &  SAR \cite{li2019show} + DA & 91.9 & 84.8 & 93.7 & 88.9  \\
         &  SCATTER \cite{litman2020scatter} + DA  & 92.0 & 84.6 & 93.6 & 89.2 \\
         \hline
         
         \multirow{3}{*}{\rotatebox[origin=c]{90}{GA}} & ASTER \cite{shi2018aster} + GA & 91.2 & 83.6 & 93.7 & 88.0 \\
         &   SAR \cite{li2019show} + GA & 91.8 & 84.8 & 93.8 & 88.4 \\
         &  SCATTER \cite{litman2020scatter} + GA  & 92.0 & 85.2 & 93.8 & 88.7 \\
         \hline
         
        \multirow{3}{*}{\rotatebox[origin=c]{90}{Augmnt.}} & Luo \etal \cite{luo2020learn} & 92.5 & 86.0 & 95.6 & 90.8 \\
        & AFDM \cite{bhunia2019handwriting} & 91.2 & 83.6 & 95.2 & 88.2 \\
        & Luo \etal + AFDM \cite{bhunia2019handwriting} & 92.7 & 86.7 & 96.1 & 91.3 \\
        \hline
         
         \multirow{18}{*}{\rotatebox[origin=c]{90}{Meta Learning based Adaptation}} & ASTER \cite{shi2018aster} Baseline & \blue{90.3} & \blue{81.3} & \blue{93.6} & \blue{87.4} \\
         & ASTER \cite{shi2018aster}  + MAML  & 93.0 & 87.1 & 96.3 & 91.9 \\
         & ASTER \cite{shi2018aster}  + MAML-FO  & 92.9 & 86.9 & 96.2 & 91.6 \\
         & ASTER \cite{shi2018aster}  + MetaSGD  & 91.1 & 83.4 & 93.7 & 88.0 \\
         & ASTER \cite{shi2018aster}  + ANIL  & 93.0 & 87.0 & 96.2 & 91.7 \\
         & ASTER \cite{shi2018aster} + \textbf{Ours (MetaHTR)} & \red{94.1} & \red{89.2} & \red{96.5} & \red{93.4} \\
         
          \cline{2-6}
          & SAR \cite{li2019show}  Baseline  & \blue{91.6} & \blue{84.4} & \blue{93.8} & \blue{88.7} \\
         & SAR \cite{li2019show}  + MAML  & 94.1 & 89.1 & 96.4 & 92.4 \\
         & SAR \cite{li2019show} + MAML-FO  & 94.0 & 88.8 & 96.3 & 92.2 \\
         & SAR \cite{li2019show}  + MetaSGD  & 91.8 & 84.9 & 93.9 & 88.9 \\
         & SAR \cite{li2019show}  + ANIL  & 94.0 & 88.9 & 96.3 & 92.3 \\
         & SAR \cite{li2019show} + \textbf{Ours (MetaHTR)} & \red{94.8} & \red{91.5} & \red{96.5} & \red{93.7} \\
          \cline{2-6}
        & SCATTER \cite{litman2020scatter}  Baseline  & \blue{91.7} & \blue{84.6} & \blue{93.6} & \blue{88.8} \\
         & SCATTER \cite{litman2020scatter}   + MAML  & 94.1 & 89.3 & 96.4 & 92.5 \\
         & SCATTER \cite{litman2020scatter}   + MAML-FO  & 94.0 & 88.9 & 96.3 & 92.3 \\
         & SCATTER \cite{litman2020scatter}   + MetaSGD  & 92.0 & 85.2 & 93.9 & 89.1 \\
         & SCATTER \cite{litman2020scatter}   + ANIL  & 94.1 & 89.1 & 96.4 & 92.4 \\
         & SCATTER \cite{litman2020scatter} + \textbf{Ours (MetaHTR)} & \red{94.8} & \red{91.6} & \red{96.6} & \red{93.9} \\
         \hline
    \end{tabular}}
    \vspace{0.1cm} 
    \label{tab2}
\end{table}

\vspace{0.1cm}
\subsection{Result Analysis and Discussion}\label{analysis}
\vspace{-0.1cm} 

The unconstrained WRA on IAM is used to cite any performance gap for describing rest of the paper, unless mentioned otherwise. In Table \ref{tab1}, we compare our MetaHTR framework with corresponding state-of-the-art (SOTA) baselines \cite{shi2018aster, li2019show, litman2020scatter} and naive fine-tuning \cite{triantafillou2019meta} method. MetaHTR outperforms (Figure \ref{fig:fig3}) every SOTA baseline by a significant margin of around $5$-$7\%$.

Furthermore, we compare with \emph{five} classes of alternative approaches in Table \ref{tab2} to tackle the style variation from different writers. We observe the following: \textbf{(i) GA:} While naive-fine-tuning hardly gives any improvement, the generative approach opens the room for generating multiple images with different words by mimicking some particular writer's handwriting style from the same adaptation set as used in MetaHTR. This being followed by naive fine-tuning does improve over the baseline HTR models by $2.3\%$, but lags behind our MetaHTR framework by $5.6\%$ for ASTER baseline. We attribute this to the inconsistency of style in the generated image \cite{kang2020ganwriting} with respect to any given writer. \cut{Due to fine-tuning through synthetically generated images, the HTR model sometimes fails in case of real handwritten samples because of the inherent domain gap \cite{guo2020face}.} Fine-tuning via synthetically generated images hurts the HTR model performance on real handwritten samples due to the inherent domain gap \cite{guo2020face}. Furthermore, style conditioned handwritten image generation involving a separate cumbersome network make it computationally more expensive.
\textbf{(ii) DG:} Although DG approach \cite{guo2020face} improves the performance  for unseen writers compared to the baseline models, it still lags behind our MetaHTR framework by $4.6\%, 5.8\%, 5.7\%$ with respect to our three baselines due to obvious reasons of not exploiting writer specific few-shot labelled data during inference. Moreover, this could be a very straight forward alternative for cases where we do not have any access to specific writer's samples, but can enrich the model with writer-specific data distribution via episodic training \cite{guo2020face}, to learn a common knowledge across writers for better performance than baseline models. \textbf{(iii) Augmnt:}  Augmentation based approach improves the performance on top of baseline models by incorporating synthetic learnable deformations, both in image space \cite{luo2020learn} and feature space \cite{bhunia2019handwriting}. Having no option of using specific writer's handwritten examples, this falls inferior to our proposed method as well. \textbf{(iv) DA:}  The performance of DA is found to be limited due to scarce adaptation data scenario and alleged instability of adversarial training. \textbf{(v) Meta Learning based Adaptation:}  Our close competitors are gradient-based meta-learning alternatives \cite{hospedales2020meta}. Out all of these, MAML \cite{finn2017model} scores quite close to ours, yet lags by $2.1\%$, $2.4\%$ $2.3\%$ for ASTER, SAR and SCATTER respectively. Although first-order approximation of MAML (MAML-FO) is computationally simpler, unconstrained WRA drops by $0.2\%$ compared to MAML on ASTER baseline. We want to emphasise that our model needs second-order gradient computation \cite{finn2017model} in the outer loop process as $g_\gamma$ is related to $\mathcal{L}^{outer}$ through inner loop update. Meta-SGD \cite{li2017meta} needs double the number of parameters than MAML as it meta-learns learning rate value for every parameter. To our surprise however, it performs lower than MAML fitted on top of our baseline text recognition models. Probably, the need of extensive parameter updates leads towards over-fitting by the meta-learning process, thus failing to generalise. 
In contrast, we use layer-wise learnable learning-rate in MetaHTR which is computationally way less expensive and provides better generalisation. Although computationally cheaper, ANIL \cite{raghu2019rapid} is still inferior to MAML baseline. We attribute the superiority of our method over other  gradient based meta-learning algorithms for sequence recognition task (e.g. HTR) to two main factors: learnable character-instance specific weighting mechanism for inner-loop loss and re-designing layer-wise learnable learning rate.

\vspace{-.2cm}
\setlength{\textfloatsep}{0pt} 
\begin{figure}[!hbt]
\begin{center}
  \includegraphics[height=2.5cm, width=\linewidth]{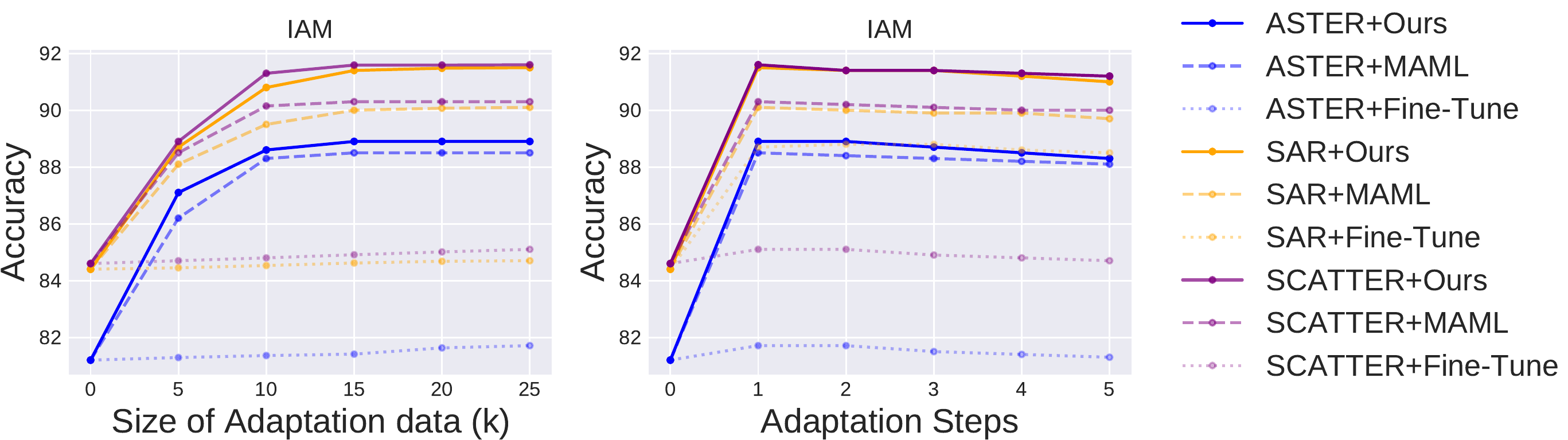}
\end{center}
\vspace{-.15in}
  \caption{Unconstrained WRA of MetaHTR with varying adaptation set-size (k), and adaptation steps (best viewed when zoomed).
}
\label{fig:4}
\vspace{-.17in}
\end{figure}

\subsection{Ablation Study}\label{ablation}
\begin{figure}[!hbt]
\begin{center}
  \includegraphics[width=\linewidth]{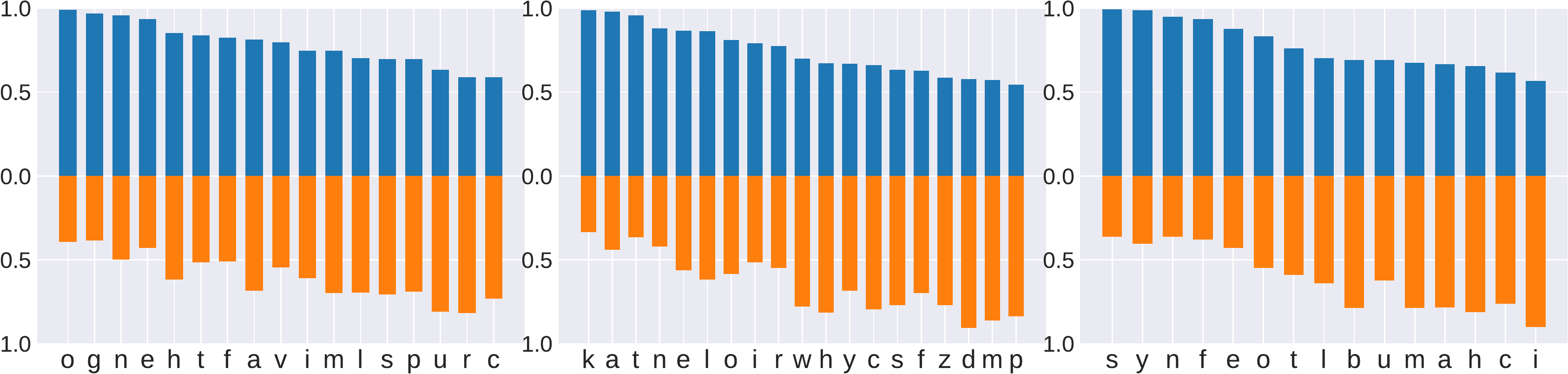}
\end{center}
\vspace{-.40cm}
  \caption{Different sets of characters (support set) of any writer (three writers shown here) show varying recognition accuracy (upper blue), thus signifying different levels of discrepancy against MetaHTR's initialization parameter. Characters having lower accuracy \emph{mostly} get higher weights (lower orange) in the inner-loop adaptation process, and vice versa. Note: X-axis for different writers is sorted in terms of recognition accuracy (Best if zoomed).
  }
\vspace{-0.3cm}
\label{fig:fig5}
\end{figure}

\vspace{-0.18cm}
\keypoint{\emph{[i] Significance of learnable $\mathbf{\gamma_t}$:}}(a) To show the efficacy of the learnable instance specific weight for character specific loss, we remove $g_\gamma$ and use simple mean cross-entropy loss (Eqn. \ref{eqn_ce}) for inner-loop update. By doing this, the performance drops by $1.9\%$, $2.2\%$ and $2.1\%$ for ASTER, SAR and SCATTER, respectively on IAM dataset.  (b) Next, we get deeper to verify whether different characters of a same writer really shows discrepancy \cite{baik2020learning} or not. For that, we evaluate character specific accuracy using our Meta-HTR model with learned initialization parameter \cite{finn2017model}. HMM-based Viterbi Forced alignment is used to locate and crop out every character from word images in a cost-effective way. From the Figure \ref{fig:fig5}, it is qualitatively evident that there exist significant variation in terms of recognition accuracy across different characters -- signifying that a few characters are harder to adapt or recognise, than others due to wide style discrepancy. For further analysis, we plot the average (for support set) character instance specific weight predicted by our meta-learned model with respect to a particular writer, for which the result is fairly consistent compared to character recognition result. This indicates that those characters which obtain low recognition accuracy, mostly receive higher weights in the inner loop loss calculation, and vice-versa, which strongly supports our intuition. (c) Furthermore, we try to explore individual gradient $\nabla_{\phi}\mathcal{L}_{ce}^t$ coming from every $t$-th character prediction of attentional decoder without concatenating it with the mean gradient $\nabla_{\phi}\mathcal{L}_{c} $ for $\gamma_t$ calculation (Eqn. \ref{eqn_gamma}).  However, the performance drop (by $1.9\%$) using ASTER baseline implies that character-instance specific gradient along with mean gradient, provides more context to judge the character wise style discrepancy with respect to the initialisation parameters.

\noindent \textbf{\emph{[ii] Layer-wise learnable learning rate:}} To analyse the contribution of learnable layer-wise learning rate mechanism, we replace it with a fixed inner-loop learning rate of $0.001$ (optimised) keeping rest of the design same. This leads to a drop of $0.8\%$, $0.6\%$ and $0.6\%$ using respective three baselines, thus justifying its contribution. Furthermore, MetaSGD \cite{li2017meta} is nearly 2.5x times computationally expensive than MAML on our HTR baselines during inference. Although computational overhead rises due to our module $g_\gamma$, the performance gain of nearly $2\%$ overshadows additional 0.2x inference time compared to MAML under similar setup. 
\noindent \textbf{\emph{[iii]  Size of adaptation data:}} A few examples are enough to achieve instant adaptation as suggested by many existing few-shot works \cite{qian2019domain, choi2020scene}. Here, we vary the size of adaptation set (k) in Figure \ref{fig:4} to discover the effect on recognition accuracy. The performance nearly saturates between 10-20 samples, thus justifying the few-shot design. Moreover, our MetaHTR tends towards saturation with slightly less samples compared to MAML under same setup. 
\noindent \textbf{\emph{[iv]  Number of adaption steps:}} The number of adaptation steps during inference is varied in Figure \ref{fig:4}.
\cut{We also vary the number of adaptation steps during inference in Figure \ref{fig:4}, and analyse its effect.} In summary, just a single gradient step update, used in most of our experiments, shows highest performance gain. On the contrary, more updates sometimes showed diminishing results that contradicts the tendency reported in \cite{finn2017model}. The reasons might be that inner loop concentrates on unnecessary style details, thus forgetting the generic prior knowledge learned.

\vspace{-.3cm}
\section{Conclusion}
\vspace{-.15cm}
    In this paper, we proposed a novel writer adaptive offline handwritten text recognition framework which aims to fully utilise the additional writer specific information available at test time. We employ an extension of Model Agnostic meta-learning (MAML) algorithm to train our writer adaptive HTR network that can quickly adapt its parameters according to the writer's handwritten text. The proposed framework is applied to three existing text recognition models without changing its architecture, and shows consistently improved performance on multiple handwritten benchmarks datasets.

{\small
\bibliographystyle{ieee_fullname}
\bibliography{Original_egbib}
}




 
 

\end{document}


\onecolumn{
\title{Supplementary material for \\ MetaHTR: Towards \emph{Writer-Adaptive} Handwritten Text Recognition}

\author{Ayan Kumar Bhunia\textsuperscript{1} \hspace{.2cm} Shuvozit Ghose\thanks{Interned with SketchX} \hspace{.2cm} Amandeep Kumar\footnotemark[1]\hspace{.2cm}   Pinaki Nath Chowdhury\textsuperscript{1,2}\hspace{.2cm}  \\ Aneeshan Sain \textsuperscript{1,2}\hspace{.2cm} Yi-Zhe Song\textsuperscript{1,2} \\
\textsuperscript{1} SketchX, CVSSP, University of Surrey, United Kingdom. 
\\
\textsuperscript{2} iFlyTek-Surrey Joint Research Centre
on Artificial Intelligence. 
\\{\tt\small \{a.bhunia, p.chowdhury, a.sain, y.song\}@surrey.ac.uk}.  \\ {\tt\small\{shuvozit.ghose,  kumar.amandeep015\}@gmail.com.}}}

\maketitle

 \appendix

\renewcommand{\thesubsection}{\Alph{subsection}}
\setcounter{figure}{0}
\setcounter{table}{0}


\section{More clarity on learnable  instance specific weights and novelty behind it}
 First, it is important to note that the instance-specific weighting mechanism is an \textit{integral} part of the meta-learning framework. It \emph{can not} be applied to learning of baseline models without meta-learning, as no fixed weight labels exist to supervise the learning process. As such, we let the model meta-learn instance-specific weights for the character-wise cross-entropy loss during adaptation. That is, the model itself needs to implicitly discover the weights, and adaptively assign weights to different characters so to prioritise learning from more discrepant characters.

\section{Clarification on marginal improvement via fine-tuning} 
 
This is not a surprise -- fine-tuning \cite{triantafillou2019meta} requires significantly more data and hundreds of gradient updates, yet gradient-based meta-learning (e.g,. MAML) does it with instantaneously with one single gradient update and with minimal data ($\leq16$ for ours). The fact that our meta-learning approach can achieve superior performance despite the disadvantages is in fact a good reflection on the effectiveness of our method.

\section{More details on experimental setup and analysis:}

(i)  $W_S/W_T$ denotes writer-set, while $D_S/D_T$ denotes the set of writer specific word-images paired with annotations.
\vspace{0.1cm}

(ii) As calculating gradients with respect to all the model's parameters is quite cumbersome, we calculate gradient of $t$-th character specific cross-entropy loss with respect to final classification layer (parameter $\phi$) as $\nabla_{\phi} \mathcal{L}^{t}_{ce}(\theta)$. This statement refers to the inner loop update that involves calculating character specific weights. In our formulation, inside the inner loop (Eq. 5-7), gradient needs to be calculated $T$ times where $T$ represents  the character sequence length of word image. Hence, considering the gradient of the whole network would have been rather cumbersome. Instead, the gradient only w.r.t. final classification layer seems to give reasonable results at a much better latency. 

\vspace{0.1cm}

(iii) As we consider $k$=$16$ in the adaptation set, we want to ensure that evaluation is done on writers having at least 16 word images to reduce the randomness.  
\vspace{0.1cm}

(iv) As adaptation set is not defined in the dataset, we randomly sample 16 images for adaptation set and rest are used for evaluation. We ensured this split to remain constant w.r.t. every writer, for every competitor for fair evaluation. Because of this evaluation protocol, our numbers for ASTER are a bit different from those reported in \cite{bhunia2019handwriting}.

{\small
\bibliographystyle{ieee_fullname}
\bibliography{Original_egbib}
}